\documentclass[twoside,11pt]{article}
\usepackage[mathlines]{lineno}
\usepackage{amsmath}
\usepackage{mathrsfs}
\usepackage{dsfont}
\usepackage[utf8]{inputenc} 
\usepackage{threeparttable}
\usepackage{float}
\usepackage{enumitem}
\usepackage{amsmath,amssymb}
\usepackage{latexsym}
\usepackage{longtable}
\usepackage{booktabs}
\usepackage{epstopdf}
\usepackage{tabularx}
\usepackage{tikz}
\usetikzlibrary{shapes.geometric, arrows}
\tikzstyle{startstop} = [rectangle, rounded corners, minimum width=3cm, minimum height=1cm, text centered, draw=black, fill=red!30]
\tikzstyle{process}   = [rectangle, minimum width=3cm, minimum height=1cm, text centered, draw=black, fill=blue!30]
\tikzstyle{io}        = [trapezium, trapezium left angle=70, trapezium right angle=110, minimum width=3cm, minimum height=1cm, text centered, draw=black, fill=green!30]
\tikzstyle{arrow}     = [thick,->,>=stealth]
\usepackage[numbers]{natbib}
\usepackage{geometry}  
\usepackage[active]{srcltx}
\usepackage{graphicx}
\usepackage{mathtools}         
  \mathtoolsset{showonlyrefs}
\usepackage{booktabs}
\usepackage{multirow}
\usepackage{siunitx}
\usepackage{xurl} 
\usepackage{amsmath}
\usepackage{setspace}
\usepackage{algorithm}         

\usepackage{amsfonts}
\usepackage{algpseudocode}
\usepackage{subcaption} 
\usepackage{natbib}
\setcitestyle{square}
\usepackage[
    bookmarks=true,
    bookmarksnumbered=true,
    colorlinks=true,
    pdfstartview=FitV,
    linkcolor=blue,
    citecolor=blue,
    urlcolor=blue
]{hyperref}

\usepackage{fancyhdr}
\usepackage{lipsum}
\usepackage{listings}
\usetikzlibrary{decorations.pathreplacing}
\usepackage{xcolor}

\definecolor{codegreen}{rgb}{0,0.6,0}
\definecolor{codegray} {rgb}{0.5,0.5,0.5}
\definecolor{codepurple}{rgb}{0.58,0,0.82}
\definecolor{backcolour}{rgb}{0.95,0.95,0.92}
\lstdefinestyle{mystyle}{
    backgroundcolor=\color{backcolour},
    commentstyle=\color{codegreen},
    keywordstyle=\color{magenta},
    numberstyle=\tiny\color{codegray},
    stringstyle=\color{codepurple},
    basicstyle=\footnotesize\ttfamily,
    breakatwhitespace=false,
    breaklines=true,
    captionpos=b,
    keepspaces=true,
    numbers=left,
    numbersep=5pt,
    showspaces=false,
    showstringspaces=false,
    showtabs=false,
    tabsize=2
}
\providecommand{\U}[1]{\protect\rule{.1in}{.1in}}

\usepackage{titlesec}
\titleformat{\section}{\normalfont\Large\bfseries}{\thesection.}{1em}{}

\begin{document}
\title{CopulaSMOTE: A Copula-Based Oversampling Approach for Imbalanced Classification in Diabetes Prediction}

\author{Agnideep Aich\textsuperscript{a}, Md Monzur Murshed\textsuperscript{b}, Bruce Wade\textsuperscript{c} and Sameera Hewage\textsuperscript{d}\thanks{CONTACT Sameera Hewage. Email: sameerahewage@suu.edu}}

\date{%
\textsuperscript{a}{Department of Emergency Medicine, Stanford University School of Medicine, Stanford, CA 94305, USA.} \\%
\textsuperscript{b}{Department of Mathematics and Statistics, Minnesota State University, Mankato, MN 56001, USA.} \\%
\textsuperscript{c}{Department of Mathematics, University of Louisiana at Lafayette, Lafayette, LA 70504, USA.} \\%
\textsuperscript{d}{Department of Mathematics, Southern Utah University, Cedar City, UT 84720, USA.}\\[2ex]%
}

\maketitle
\begin{abstract}
Class imbalance remains a practical obstacle in the development of clinical prediction models for conditions such as diabetes mellitus, where the number of confirmed cases is often much smaller than the number of controls. The Synthetic Minority Over-sampling Technique (SMOTE) and its variants are widely used to address this imbalance, but they generate synthetic observations through local interpolation in feature space and do not explicitly model the joint dependence structure of the minority class. To address this challenge, our study 
introduces a copula-based data augmentation approach that estimates the minority-class
dependence structure when generating synthetic samples and integrates with standard
machine learning techniques. Specifically, we employ truncated vine copulas to represent
multivariate dependence through a sequence of bivariate building blocks. We evaluate the proposed approach on three public diabetes datasets, namely the Pima Indians Diabetes dataset, the Iraqi Diabetes dataset, and the CDC BRFSS 2015 Diabetes Health Indicators dataset, which together cover a range of sample sizes, dimensionalities, and imbalance regimes. For each dataset, five resampling strategies are compared across five classifiers using a 5 by 2 cross validation protocol with Dietterich's paired t test. Our findings suggest that CopulaSMOTE can improve minority-class recovery in larger tabular diabetes datasets, particularly the CDC BRFSS dataset, but its advantages depend on the classifier and evaluation metric.

\end{abstract}



\section{Introduction}

Diabetes mellitus is a chronic metabolic disorder of growing global burden.
According to the International Diabetes Federation, approximately 537 million
adults were living with diabetes in 2021, a figure projected to reach 783
million by 2045, with a substantial proportion of cases remaining undiagnosed
at the time of screening \cite{IDF2021}. In the United States alone, more than
one in five diabetic adults is undiagnosed \cite{CDC2022diabetes}. Undetected
diabetes is associated with prolonged hyperglycemia, progression to
microvascular and macrovascular complications, and substantially higher
long-term healthcare costs \cite{zheng2018global}. These considerations
motivate sustained interest in machine learning approaches to diabetes risk
prediction from routinely collected clinical and survey data.

Studies using the Pima Indians Diabetes dataset \cite{smith1988using}, the
CDC Behavioral Risk Factor Surveillance System (BRFSS) \cite{cdc2015brfss},
and other registries have demonstrated that classifiers can identify high-risk
individuals from features such as plasma glucose, BMI, insulin levels, age,
and hypertension status \cite{kavakiotis2017machine, zou2018predicting,
tigga2020prediction}. A persistent challenge in this setting is
\emph{class imbalance}. In population surveys such as the CDC BRFSS, the
diabetes-to-non-diabetes ratio can be as low as 1:8 \cite{teboul2021diabetes}.
Standard classifiers trained on such data exhibit poor sensitivity for the
positive class \cite{johnson2019survey}, which is the clinically critical
outcome. A missed diagnosis directly delays treatment and accelerates
disease progression.

The Synthetic Minority Over-sampling Technique (SMOTE) \cite{chawla2002smote}
and its variants, including Borderline-SMOTE \cite{han2005borderline}, ADASYN
\cite{he2008adasyn}, SMOTE-ENN \cite{batista2004study}, SMOTE-Tomek
\cite{tomek1976two}, and K-Means SMOTE \cite{douzas2018improving}, address
this imbalance by interpolating between neighboring minority observations in
feature space. Despite their broad utility, these methods do not explicitly
model the global joint dependence structure of the minority class.
This limitation is consequential for diabetes data, where risk factors such
as glucose tolerance, insulin resistance, BMI, and blood pressure are not
marginally independent. They reflect the shared pathophysiology of metabolic
syndrome \cite{alberti2006metabolic}. Synthetic samples generated without
accounting for these dependencies risk being metabolically implausible and
may introduce noise into the training distribution. Deep generative models
such as GANs and VAEs can capture complex joint structure, but they require
large training samples, involve substantial computational overhead, and offer
limited interpretability in clinical settings \cite{frid2018gan, xu2019modeling,
kingma2013auto}. These are practical disadvantages in the small-to-moderate
sample settings characteristic of most diabetes datasets.

Copula models \cite{sklar1959fonctions, nelsen2006introduction} provide a
principled alternative. They decompose the joint distribution into marginal
distributions and a dependence structure, allowing each to be modeled
separately. Among copula families, vine copulas
\cite{bedford2001probability, aas2009pair} are particularly well-suited to
moderate-dimensional tabular data. They represent multivariate dependence as
sequences of bivariate building blocks, each assigned an independently
selected copula family. This allows heterogeneous dependence patterns, such
as those observed among diabetes risk factors, to be captured flexibly
and at scale.

In this paper, we introduce \textbf{CopulaSMOTE}, a vine copula-based
oversampling framework for imbalanced diabetes classification. The method
fits a vine copula to the minority class, generates synthetic samples in the
copula domain, and maps them back to the original feature space via the
fitted marginals. We evaluate it on three public diabetes datasets spanning
a range of sample sizes, dimensionalities, and imbalance regimes. We compare CopulaSMOTE with four baseline resampling strategies across five classifiers using a
$5\times2$ cross-validation protocol with Dietterich's paired $t$-test
\cite{dietterich1998approximate}.

The main contributions of this study are as follows:
\begin{enumerate}[leftmargin=*, label=(\roman*)]
    \item We propose CopulaSMOTE, a synthetic oversampling method that estimates minority-class
dependence using a truncated vine copula and preserves empirical marginal support through
an empirical inverse CDF transformation.
    \item We evaluate the method on the Pima Indians Diabetes dataset,
          the Iraqi Diabetes dataset, and the CDC BRFSS 2015 Diabetes
          Health Indicators dataset.
    \item We provide empirical comparisons showing that CopulaSMOTE improves F1 score and
minority-class recall most clearly in the CDC BRFSS setting, while AUC and PR AUC remain classifier dependent.
\end{enumerate}
\newpage
The remainder of the paper is organized as follows. Section~\ref{sec:related}
reviews related work. Section~\ref{sec3} introduces the methodology. Section~\ref{sec4}
describes the experimental setup. Section~\ref{sec5} presents the results.
Section~\ref{sec6} concludes.


\section{Related Work}\label{sec:related}

\subsection{Machine Learning for Diabetes Prediction}

The application of machine learning to diabetes risk prediction has expanded
substantially over the past two decades, driven by the availability of large
clinical and survey-based datasets and the demonstrated limitations of
rule-based screening criteria. Comprehensive reviews by Kavakiotis et al.\ \cite{kavakiotis2017machine}
and Afsaneh et al.\ \cite{afsaneh2022review} document the progression
from early logistic regression and na\"ive Bayes classifiers to modern
ensemble methods and deep neural networks, finding that gradient boosting
and random forest consistently outperform linear models on tabular
clinical data. Early work by \cite{smith1988using} established the Pima
Indians Diabetes dataset as the canonical benchmark for binary classification
under moderate class imbalance, and it has since been used in hundreds of
comparative studies. Reported AUC values on Pima range widely, from 0.75
for simple logistic regression to 0.87 for tuned ensemble methods, with the
spread attributable largely to differences in preprocessing and resampling
strategy rather than classifier architecture
\cite{zou2018predicting, tigga2020prediction}.

More recent work has shifted toward larger tabular datasets with richer predictor sets.
On the CDC BRFSS data \cite{cdcdiabetes}, Alghamdi et al.\ \cite{alghamdi2017predicting}
showed that ensemble methods with SMOTE-based oversampling can identify
high-risk individuals from self-reported demographic and health indicators,
with gradient boosting and random forest offering the strongest performance. A recurring finding
across this literature is that class imbalance is not merely a nuisance
but a systematic source of degraded minority-class recall: standard
classifiers trained without resampling consistently achieve high specificity
and low sensitivity, regardless of classifier choice
\cite{maniruzzaman2017comparative, johnson2019survey}. This pattern is
especially pronounced in population surveys such as the CDC BRFSS, where
the diabetes-to-non-diabetes ratio can be as low as 1:8
\cite{teboul2021diabetes}, and has motivated a sustained body of work on
oversampling as a preprocessing step for clinical prediction models
\cite{khanam2021comparison}.

Despite this progress, two limitations persist in the existing literature.
First, the majority of studies treat oversampling as a black-box
preprocessing step and do not examine whether the synthetic observations
are statistically plausible with respect to the joint dependence structure
of the minority class. Second, comparative evaluations typically focus on
a single dataset and a limited set of classifiers, making it difficult
to assess how the benefits of a given resampling strategy vary with
dataset dimensionality and minority-class size. The present work addresses
both limitations directly.

\subsection{Copula-Based Synthetic Data Generation}

Copula models decompose a multivariate distribution into marginal distributions
and a dependence structure through Sklar's theorem
\cite{sklar1959fonctions, nelsen2006introduction}. For a $d$-dimensional joint
distribution $F$ with continuous marginals
$F_1, \dots, F_d$, there exists a unique copula
$C : [0,1]^d \rightarrow [0,1]$ such that
\begin{equation}
    F(x_1, \dots, x_d)
    =
    C\bigl(F_1(x_1), \dots, F_d(x_d)\bigr).
    \label{eq:sklar}
\end{equation}

Equivalently, the joint density can be factorized as
\begin{equation}
    f(x_1, \dots, x_d)
    =
    c\bigl(F_1(x_1), \dots, F_d(x_d)\bigr)
    \prod_{j=1}^{d} f_j(x_j),
    \label{eq:sklar_density}
\end{equation}
where $c$ denotes the copula density and $f_j$ are the marginal densities.
This separation is particularly attractive for clinical tabular data, where
marginal distributions are often heterogeneous and dependence relationships
may contain clinically meaningful structure.

\begin{figure}[htbp]
\centering
\begin{tikzpicture}[
    box/.style={rectangle, rounded corners=4pt, draw=black,
                minimum width=2.8cm, minimum height=0.85cm,
                text centered, font=\small},
    arr/.style={->, thick, >=stealth}
]

\node[box, fill=blue!15]   (joint) at (3.0, 0)
    {Joint distribution $F(x_1,\dots,x_d)$};

\node[box, fill=green!15]  (marg)  at (0.5, -2.0)
    {Marginals $F_1,\dots,F_d$};

\node[box, fill=orange!15] (cop)   at (5.5, -2.0)
    {Copula $C(u_1,\dots,u_d)$};

\node[box, fill=gray!12, text width=5.8cm] (synth) at (3.0, -4.0)
    {$F(\mathbf{x}) = C\!\left(F_1(x_1),\dots,F_d(x_d)\right)$};

\draw[arr] (joint) -- (marg)
    node[midway, above left, font=\footnotesize]{marginal behaviour};
\draw[arr] (joint) -- (cop)
    node[midway, above right, font=\footnotesize]{dependence structure};
\draw[arr] (marg)  -- (synth);
\draw[arr] (cop)   -- (synth);

\end{tikzpicture}
\caption{Sklar's theorem decomposes a joint distribution into its marginal
distributions and a copula capturing the dependence structure. The two
components can be modeled and sampled independently, which is the key
property exploited by CopulaSMOTE.}
\label{fig:sklar_decomp}
\end{figure}

Several recent studies have explored copula-based synthetic data generation.
Restrepo et al.\ \cite{restrepo2023nonparametric} proposed a nonparametric
copula-based oversampling method that outperformed SMOTE by better preserving
multivariate dependence relationships. Wan et al.\ \cite{wan2019sync}
introduced the SynC framework based on Gaussian copulas for synthetic
population generation, while Houssou et al.\ \cite{houssou2022generation}
demonstrated that copula-generated data can preserve important multivariate
statistical characteristics in biomedical datasets.

\subsection{Vine Copulas for High-Dimensional Dependence Modeling}

Vine copulas \cite{bedford2001probability, bedford2002vines} extend copula
modeling to high-dimensional settings by decomposing a joint distribution into
a sequence of bivariate conditional copulas arranged through a regular vine
structure. Estimation procedures for pair-copula constructions were developed
by \cite{aas2009pair}, and comprehensive methodological treatments are provided
by \cite{czado2019analyzing}.

For a $d$-dimensional random vector, a regular vine (R-vine) represents the
joint copula density as a cascade of conditional bivariate copulas. The full
density decomposition can be written as
\begin{equation}
    f(x_1, \dots, x_d)
    =
    \prod_{j=1}^{d} f_j(x_j)
    \prod_{m=1}^{d-1}
    \prod_{e \in T_m}
    c_{j(e),k(e)\mid D(e)}
    \!\left(
    F_{j(e)\mid D(e)}(x_{j(e)}\mid x_{D(e)}),
    F_{k(e)\mid D(e)}(x_{k(e)}\mid x_{D(e)})
    \right).
    \label{eq:vine}
\end{equation}
where $T_1, \dots, T_{d-1}$ denote the sequence of vine trees,
$c_{j(e),k(e)\mid D(e)}$ is the pair-copula associated with edge $e$, and
$F_{j(e)\mid D(e)}$ represents the conditional distribution function of
variable $j(e)$ given the conditioning set $D(e)$
\cite{aas2009pair, czado2019analyzing}.

\begin{figure}[htbp]
\centering
\begin{tikzpicture}[
    vnode/.style={circle, draw=black, fill=blue!15,
                  minimum size=0.75cm, font=\small, inner sep=2pt},
    enode/.style={rectangle, rounded corners=3pt, draw=black,
                  fill=orange!20, minimum width=1.6cm,
                  minimum height=0.65cm, font=\scriptsize,
                  text centered, inner sep=3pt},
    arr/.style={-, thick}
]

\node[font=\small\bfseries] at (-0.4, 0.6) {$T_1$};
\node[vnode] (n1) at (0.0, 0)  {$1$};
\node[vnode] (n2) at (2.2, 0)  {$2$};
\node[vnode] (n3) at (4.4, 0)  {$3$};
\node[vnode] (n4) at (6.6, 0)  {$4$};

\draw[arr] (n1)--(n2) node[midway, above, font=\scriptsize]{$c_{12}$};
\draw[arr] (n2)--(n3) node[midway, above, font=\scriptsize]{$c_{23}$};
\draw[arr] (n3)--(n4) node[midway, above, font=\scriptsize]{$c_{34}$};

\node[font=\small\bfseries] at (-0.4, -2.0) {$T_2$};
\node[enode] (e12) at (1.4, -2.0) {$1,2$};
\node[enode] (e23) at (3.8, -2.0) {$2,3$};
\node[enode] (e34) at (6.2, -2.0) {$3,4$};

\draw[arr] (e12)--(e23) node[midway, above, font=\scriptsize]{$c_{13|2}$};
\draw[arr] (e23)--(e34) node[midway, above, font=\scriptsize]{$c_{24|3}$};

\node[font=\small\bfseries] at (-0.4, -4.0) {$T_3$};
\node[enode] (e13) at (2.6, -4.0) {$1,3|2$};
\node[enode] (e24) at (5.0, -4.0) {$2,4|3$};

\draw[arr] (e13)--(e24) node[midway, above, font=\scriptsize]{$c_{14|23}$};

\draw[dashed, gray!50] (1.4,  -0.38) -- (1.4,  -1.68);
\draw[dashed, gray!50] (3.8,  -0.38) -- (3.8,  -1.68);
\draw[dashed, gray!50] (6.2,  -0.38) -- (6.2,  -1.68);
\draw[dashed, gray!50] (2.6,  -2.38) -- (2.6,  -3.68);
\draw[dashed, gray!50] (5.0,  -2.38) -- (5.0,  -3.68);

\end{tikzpicture}
\caption{A D-vine structure for four variables illustrating the three trees
$T_1$, $T_2$, $T_3$ of a regular vine copula. Each edge corresponds to a
bivariate copula: unconditional pairs in $T_1$ (e.g.\ $c_{12}$, $c_{23}$,
$c_{34}$), and conditional pairs in subsequent trees (e.g.\ $c_{13|2}$,
$c_{24|3}$, $c_{14|23}$). Each bivariate copula can be drawn from a
different parametric family, allowing heterogeneous dependence structures
to be captured across the feature space.}
\label{fig:vine_tree}
\end{figure}
An advantage of vine copulas is that different pairwise components can be modeled using
different copula families. This allows heterogeneous dependence patterns to be represented
within a single modeling framework. This property is
particularly useful for clinical and epidemiological data, where relationships
between variables may exhibit asymmetric dependence or tail-dependent behaviour.
For example, dependence between glucose and BMI may differ substantially from
dependence between age and physical activity, making a single global copula
family overly restrictive.

Despite their flexibility and strong empirical performance in dependence
modeling, vine copulas have seen limited use in class-imbalance learning and
synthetic minority oversampling. Existing work primarily focuses on density
estimation, financial risk modeling, or synthetic population generation rather
than supervised classification under severe imbalance. This gap motivates the
development of CopulaSMOTE, which uses vine copulas to generate synthetic
minority observations while approximating the minority-class dependence structure.

\subsection{Positioning of the Present Work}

Table~\ref{tab:comparison} summarizes the key properties of competing
approaches relative to CopulaSMOTE.

\begin{table}[H]
\centering
\scriptsize
\begin{threeparttable}
\caption{Comparison of oversampling strategies across key properties.}
\label{tab:comparison}
\begin{tabularx}{\textwidth}{lXXXXX}
\toprule
\textbf{Method} & \textbf{Models dependence} & \textbf{High-dim.\ scalable} & \textbf{Small-sample feasible} & \textbf{Interpretable} & \textbf{Marginal-preserving} \\
\midrule
SMOTE (variants) & No      & Partial & Yes & Yes & Partial \\
GAN / CTGAN      & Yes     & Yes     & No  & No  & Partial \\
VAE / CVAE       & Yes     & Yes     & No  & No  & Partial \\
Gaussian copula  & Partial & Partial & Yes & Yes & Yes     \\
\textbf{CopulaSMOTE}\tnote{a} & \textbf{Approximate} & \textbf{Yes} & \textbf{Yes} & \textbf{Yes} & \textbf{Empirical support preserving} \\
\bottomrule
\end{tabularx}
\begin{tablenotes}
\footnotesize
\item[a] For mixed binary, low-cardinality, and continuous predictors, CopulaSMOTE uses a continuous-extension approximation. Therefore, dependence modeling should be interpreted as approximate rather than as a full mixed-data copula likelihood.
\end{tablenotes}
\end{threeparttable}
\end{table}

\noindent The present work addresses a gap at the intersection of these strands of literature. Existing
oversampling methods usually do not explicitly model the joint dependence structure of diabetes
risk factors. Deep generative models can model complex structure, but they may be less practical
in smaller clinical samples. CopulaSMOTE offers a dependence-aware and interpretable
oversampling framework, evaluated across datasets spanning different sample sizes, dimensions,
and imbalance regimes.

\section{Methodology}\label{sec3}

This section introduces the proposed oversampling framework. The goal is to generate
synthetic minority-class observations by estimating the dependence structure of the observed
minority class with a truncated vine copula. We describe four steps: construction of
pseudo-observations, fitting of the truncated vine copula, simulation of new pseudo-observations
from the fitted vine, and back transformation to the original feature space through empirical
inverse cumulative distribution functions. Throughout, the procedure is applied strictly to the
training portion of each cross-validation fold, so the held-out test set is never used in constructing
either the copula model or the synthetic observations.

\begin{figure}[htbp]
\centering
\begin{tikzpicture}[
    obox/.style={rectangle, rounded corners=5pt, draw=black,
                 fill=blue!10, minimum width=7.0cm,
                 minimum height=1.2cm, align=center, font=\small},
    cbox/.style={rectangle, rounded corners=5pt, draw=black,
                 fill=orange!15, minimum width=7.0cm,
                 minimum height=1.2cm, align=center, font=\small},
    arr/.style={->, thick, >=stealth}
]

\node[obox] (d)  at (0,  0.0) {Minority class \quad $D_{\min} \subset \mathbb{R}^d$};
\node[cbox] (u)  at (0, -2.2) {Pseudo-observations \quad $U \in [0,1]^d$};
\node[cbox] (us) at (0, -4.4) {Synthetic pseudo-obs. \quad $U^{\text{syn}} \in [0,1]^d$};
\node[obox] (xs) at (0, -6.6) {Synthetic minority \quad $X^{\text{syn}} \subset \mathbb{R}^d$};

\draw[arr] (d)  -- (u);
\draw[arr] (u)  -- (us);
\draw[arr] (us) -- (xs);

\node[font=\scriptsize, align=left] at (2.7, -1.1)
    {rank transform: $U_{ij} = R_{ij}\,/\,(n+1)$};
\node[font=\scriptsize, align=left] at (2.2, -3.3)
    {fit \& sample: $U^{\text{syn}} \sim \widehat{C}_{\text{vine}}$};
\node[font=\scriptsize, align=left] at (1.9, -5.5)
    {inverse CDF: $F^{-1}_{j,\text{emp}}(\cdot)$};

\node[font=\footnotesize, text=gray!60, align=right] at (-5.5,  0.0) {feature space};
\node[font=\footnotesize, text=gray!60, align=right] at (-5.5, -2.2) {copula space};
\node[font=\footnotesize, text=gray!60, align=right] at (-5.5, -4.4) {copula space};
\node[font=\footnotesize, text=gray!60, align=right] at (-5.5, -6.6) {feature space};

\end{tikzpicture}
\caption{The CopulaSMOTE pseudo-observation pipeline. Minority class features
are rank-transformed to pseudo-observations on the unit hypercube $[0,1]^d$,
a truncated vine copula is fitted and sampled to produce synthetic
pseudo-observations reflecting the dependence structure estimated by the fitted vine,
and an empirical inverse CDF maps them back to the original feature space.
Blue boxes indicate steps in the original feature space; orange boxes
indicate steps in the copula domain.}
\label{fig:pseudo_obs_pipeline}
\end{figure}

\subsection{Setting and Notation}

Consider a training fold consisting of observations $(X_i, y_i)$ for $i = 1, \ldots, n$, where $X_i \in \mathbb{R}^d$ is a vector of predictors and $y_i \in \{0, 1\}$ is a binary outcome label. Let $c_{\text{min}}$ denote the minority class in the training fold, chosen as the class with the smaller number of training observations. Let $D_{\text{min}} = \{X_i : y_i = c_{\text{min}}\}$ denote the minority class feature matrix, with $n_{\text{min}}$ rows and $d$ columns. The number of synthetic observations to be generated is $n_{\text{syn}} = n_{\text{maj}} - n_{\text{min}}$, where $n_{\text{maj}}$ is the majority class count. This choice produces a perfectly balanced training fold after augmentation.

Before any resampling, the training fold is processed through the same transformations that will be applied to the held out test fold at evaluation time. Specifically, zero coded missing values in the Pima dataset are replaced by the column median using a median imputer fitted only on the training fold, and all features are standardized using a scaler fitted only on the training fold. These fitted transformations are then applied to the test fold, which keeps the resampling, classifier fitting, and evaluation steps consistent with standard practice for leakage free cross validation.

\subsection{Pseudo Observations from the Minority Class}

Vine copulas act on pseudo observations on the unit cube $(0, 1)^d$ rather than on the raw features. To map the minority class feature matrix $D_{\text{min}}$ to this scale, we use an empirical rank transformation. For each feature $j = 1, \ldots, d$ and each minority observation $i = 1, \ldots, n_{\text{min}}$, we define

$$
U_{ij} = \frac{R_{ij}}{n_{\text{min}} + 1},
$$

where $R_{ij}$ denotes the rank of the $i$-th value in the $j$-th column of $D_{\text{min}}$. The resulting matrix $U$ has entries in $(0, 1)$ and provides scale free pseudo observations of the minority class.

Binary and low-cardinality features, common in the Iraqi and CDC datasets, produce many tied values after rank transformation. Tied pseudo-observations violate the continuity assumption underlying vine copula fitting and cannot be processed directly. To handle this, we apply a small Gaussian jitter to the minority class features before rank transformation. This is the continuous extension approach that has been used in the copula and dependence literature to accommodate ties arising from discrete or mixed data \cite{denuit2005constraints, schallhorn2017d, nagler2018vine}. In practice, for each fold we draw an independent $d$ dimensional Gaussian noise matrix with component standard deviation $\varepsilon = 10^{-6}$ and add it to the standardized minority class feature matrix before computing ranks. Because the jitter magnitude is orders of magnitude smaller than the standardized feature scale, the jitter has negligible effect on the marginal distributions, and its role is restricted to breaking ties so that the rank transformation can be evaluated. We emphasize that this step is part of a practical dependence aware framework rather than a fully specified mixed data copula likelihood model, and we return to this point in the discussion.

\subsection{Truncated Vine Copula on the Minority Class}

Given the pseudo observation matrix $U$, we fit a vine copula to the minority class. Vine copulas represent a high dimensional copula density as a product of bivariate building blocks arranged along a sequence of trees \cite{bedford2001probability, bedford2002vines, aas2009pair}. Each edge of the vine corresponds to a bivariate copula that captures either an unconditional pairwise dependence or a conditional pairwise dependence given the variables on earlier trees. This construction makes it possible to capture a rich set of dependence patterns across many variables without committing to a single global parametric copula family.

For each fold, we fit a vine copula with pairwise components selected from a standard library of bivariate families, namely the independence copula, the Gaussian copula, the Student t copula, the Clayton copula, the Gumbel copula, and the Frank copula, with rotations allowed. The family for each edge is selected by the Bayesian Information Criterion (BIC), and the tree structure is determined automatically by the fitting routine. To keep the computational cost manageable across datasets of different dimensionalities, and to avoid overfitting higher order conditional dependencies on modest sized minority classes, we use a truncated vine at level $\min(3, d - 1)$, so that only the first three trees are estimated and the remaining edges are replaced by independence copulas. Truncation is a standard device in the vine copula literature for balancing flexibility with statistical stability in higher dimensional settings \cite{brechmann2012truncated}.

\subsection{Simulation and Back Transformation}

Once the truncated vine copula has been fitted to the minority class pseudo observations, we simulate $n_{\text{syn}}$ new pseudo observations from the fitted vine using its built in sampler. The simulated matrix $U^{\text{syn}}$ has entries in \((0,1)\) and reflects the dependence pattern estimated by the fitted truncated
vine copula.

To return to the original feature space, we apply an empirical inverse cumulative distribution function separately to each column of $U^{\text{syn}}$. For each feature $j$, let $v_{j,(1)} \le v_{j,(2)} \le \cdots \le v_{j,(n_{\text{min}})}$ denote the sorted values of that feature in the minority class. For each simulated pseudo observation $u_{ij}^{\text{syn}} \in (0, 1)$, the back transformed value is

$$
X_{ij}^{\text{syn}} = v_{j,\, k_{ij}}, \qquad k_{ij} = \max\!\big(1, \min(n_{\text{min}}, \lceil u_{ij}^{\text{syn}} \cdot n_{\text{min}} \rceil)\big).
$$

This nearest sorted value scheme is a fully empirical, nonparametric inverse cumulative distribution function mapping. It has the practical advantage that all values generated for a given feature are values that actually occurred in the minority class for that feature, which preserves the empirical marginal support even when a feature is binary or discrete. In particular, this mapping does not introduce interpolated values that lie outside the observed support of any feature, which we view as a desirable property for synthetic observations that will later be combined with real training data. Thus, the method should be interpreted as dependence-guided recombination of observed
minority-class feature values, rather than fully continuous generation of new marginal values.

The synthetic matrix $X^{\text{syn}}$ is labeled with the minority class and appended to the training fold, yielding a balanced augmented training set. The augmented training set is then shuffled and used to fit each of the classifiers described in the experimental section. The held out test fold is transformed only through the fitted imputer and scaler, is never augmented, and is used exclusively for evaluation.

\begin{algorithm}[htbp]
\caption{Vine Copula-Based Oversampling}
\label{alg:vine_oversampling}
\begin{algorithmic}[1]
\Require Training dataset $D = \{(X_i, y_i)\}_{i=1}^n$
\Ensure Balanced dataset $D_{\text{bal}}$

\State Identify minority class $c_{\min}$, subset $D_{\min}$, and majority subset $D_{\text{maj}}$
\State $n_{\text{syn}} \leftarrow |D_{\text{maj}}| - |D_{\min}|$

\vspace{4pt}
\State \textbf{// Copula Representation}
\State Apply Gaussian jitter to $D_{\min}$ to resolve ties
\State Compute pseudo-observations:
        $U_{ij} = \dfrac{\operatorname{rank}(X_{ij})}{n_{\min} + 1}$
        for $i = 1,\dots,n_{\min}$, $j = 1,\dots,d$
\State Fit truncated vine copula $\widehat{C}_{\text{vine}}$ to $U = (U_{ij})$
\State Sample $U^{\text{syn}} \sim \widehat{C}_{\text{vine}}$ with $n_{\text{syn}}$ draws

\vspace{4pt}
\State \textbf{// Inverse Transformation}
\For{each feature $j = 1,\dots,d$}
    \State $X^{\text{syn}}_{ij} \leftarrow F^{-1}_{j,\mathrm{emp}}\!\left(U^{\text{syn}}_{ij}\right)$
    for $i = 1,\dots,n_{\text{syn}}$
\EndFor

\vspace{4pt}
\State $D_{\text{syn}} \leftarrow \{(X^{\text{syn}}_i,\, c_{\min})\}_{i=1}^{n_{\text{syn}}}$
\State $D_{\text{bal}} \leftarrow D_{\min} \cup D_{\text{maj}} \cup D_{\text{syn}}$
\State Shuffle $D_{\text{bal}}$
\State \Return $D_{\text{bal}}$
\end{algorithmic}
\end{algorithm}

\subsection{Status of the Framework}

The procedure above is a dependence aware oversampling method in the sense that the synthetic observations inherit a joint dependence structure estimated from the minority class rather than arising solely from local geometric interpolation. It is not a fully specified discrete or mixed data copula likelihood model. In particular, the continuous extension jitter step described in Section 3.2 is used as a practical device for evaluating a vine copula on data that includes binary and low cardinality features. This is the approach that has been taken in earlier work on copula modeling with ties \cite{denuit2005constraints, schallhorn2017d, nagler2018vine}, and we adopt it here in the same practical spirit rather than as a claim of exact discrete likelihood.

\subsection{Experimental Protocol}

All experiments use a 5 by 2 cross validation protocol in which the dataset is split into two stratified halves five independent times, and the outcome of interest is the performance of each method averaged across the ten resulting folds. Within each fold, the imputer, scaler, oversampling routine, and classifier are fitted on the training half and then applied to the held out half for evaluation. The 5 by 2 design is used because it is specifically recommended by \cite{dietterich1998approximate} for comparing learning algorithms, and because it admits a paired t test with five degrees of freedom that controls the type I error rate under the conditions most relevant to the present setting.

For each method, classifier, and fold, we record two groups of performance metrics. The first group consists of overall metrics: accuracy, balanced accuracy, precision, recall, F1, AUC, and PR AUC. The second group consists of the same metrics computed with the minority class as the positive label, which we refer to as minority-focused metrics. This distinction allows consistent comparison across datasets where the minority class differs in identity. The minority focused metrics allow a consistent comparison across datasets despite the fact that the minority class corresponds to the non diabetic group in the Iraqi dataset and to the diabetic group in the Pima and CDC datasets. For hypothesis testing, we use Dietterich's 5 by 2 paired t test with five degrees of freedom. The statistic takes the form
$$
t = \frac{d_{1,1}}{\sqrt{\tfrac{1}{5}\sum_{i=1}^{5} s_i^2}}, \qquad s_i^2 = (d_{i,1} - \bar{p}_i)^2 + (d_{i,2} - \bar{p}_i)^2,
$$

where $d_{i,1}$ and $d_{i,2}$ are the per fold performance differences for iteration $i$ and $\bar{p}_i$ is their average \cite{dietterich1998approximate}. Two sided p values are reported throughout.

\section{Datasets and Experimental Setup}\label{sec4}

We evaluate the proposed framework on three publicly available diabetes datasets selected to cover a range of sample sizes, dimensionalities, and imbalance regimes.

The Pima Indians Diabetes dataset \cite{smith1988using} contains 768 observations on adult female patients of Pima Indian heritage, with eight predictors including plasma glucose, blood pressure, body mass index, and age, and a binary outcome indicating diabetes status. Five features have zero coded missing values for physiologically implausible zeros, which we set to missing values and impute by the per fold training median. The class distribution is 500 non diabetic and 268 diabetic, so the minority class is the diabetic group at approximately 34.9 percent of the sample. This dataset is included primarily as a small, low dimensional benchmark.

The Iraqi Diabetes dataset \cite{rashid}, collected from Iraqi hospital laboratory data, contains 1001 observations on eleven predictors including age, gender, urea, creatinine, HbA1c, and lipid panel components, and a three category outcome labeled non diabetic (N), pre diabetic (P), and diabetic (Y). To obtain a binary outcome consistent with the other two datasets, we drop the 53 pre diabetic observations and recode the remaining labels as 0 (non diabetic) and 1 (diabetic). After this filtering the sample size is 947, with 103 non diabetic and 844 diabetic observations. The minority class is therefore the non diabetic group at approximately 10.9 percent of the filtered sample, which reverses the usual relationship between the minority class and the clinically salient class relative to the other two datasets. The oversampling routines in our experiments identify the minority class dynamically from the training fold, so this reversal is handled automatically, but it does mean that reporting needs to be done on minority focused metrics as well as overall metrics to allow a consistent comparison across datasets. Gender is encoded as 0 for female and 1 for male, and patient identifier columns are dropped.

The CDC BRFSS 2015 Diabetes Health Indicators dataset \cite{cdcdiabetes} is a large survey based dataset with 253,680 observations and 21 predictors covering self reported health indicators, demographics, and health care access variables. The binary outcome indicates diabetes status. The class distribution is 218,334 non diabetic and 35,346 diabetic, so the minority class is the diabetic group at approximately 13.9 percent of the sample. The CDC dataset is our main setting of interest, both because it is the largest dataset and because it has the highest dimensionality among the three.

For oversampling, we compare the proposed vine copula method to four baselines. Three of the baselines are classical interpolation-based oversamplers, namely SMOTE \cite{chawla2002smote}, Borderline-SMOTE \cite{han2005borderline}, and ADASYN \cite{he2008adasyn}; hybrid methods that combine oversampling with editing are outside the scope of this comparison, which focuses on the question of whether dependence-aware generation improves upon interpolation. In rare cases where ADASYN fails to fit because of degenerate neighborhood conditions, the code falls back to SMOTE for that fold, which keeps the experiment comparable across folds. The fourth baseline is a normalizing flow, which is a generative density model that provides a deep learning comparison to the copula based approach. The flow is constructed using coupling layers with residual subnets and a standard Gaussian base distribution, is trained only on the minority class of the training fold for a fixed number of epochs, and is used to sample new minority observations that are appended to the training fold.

For classification, we use five standard classifiers, namely random forest (RF), gradient boosting (GB), extreme gradient boosting (XGB), logistic regression (LR), and a multilayer perceptron (MLP), each with its default hyperparameters aside from increasing the logistic regression iteration cap to ensure convergence. All classifiers are fitted on the oversampled training fold and evaluated on the unaugmented held out fold.

\section{Results}\label{sec5}

This section reports the empirical comparison of CopulaSMOTE against the four
baseline methods across the three datasets. Results are organized by dataset
and focus on the metrics most informative under class imbalance: F1, AUC,
balanced accuracy, and their minority-class-focused counterparts. All reported
values are mean $\pm$ standard deviation across the ten folds of the
5$\times$2 cross-validation, and all $p$-values are from Dietterich's paired
$t$-test with five degrees of freedom.

\subsection{Pima Indians Diabetes}\label{subsec:pima}

The Pima dataset is the smallest and lowest-dimensional dataset considered,
with 768 observations and eight features. Performance differences among the
five resampling strategies are narrow throughout, and only one comparison
reaches conventional statistical significance.

Table~\ref{tab:pima_iraqi_results} (Panel~A) reports mean F1 and AUC for all
method-classifier combinations. CopulaSMOTE achieves the highest numerical F1
for three of the five classifiers and the highest AUC for four, but the margins
are small relative to cross-fold variability and no consistent pattern emerges. Only one comparison reached nominal statistical significance at
\(\alpha = 0.05\): CopulaSMOTE versus SMOTE for gradient boosting F1
\((\Delta = +0.0097,\ p = 0.049)\). Because this result arises from a large
set of pairwise comparisons, it should be interpreted as exploratory rather
than definitive evidence of superiority.

\begin{figure}[htbp]
    \centering
\includegraphics[width=\textwidth,height=0.8\textheight,keepaspectratio]
    {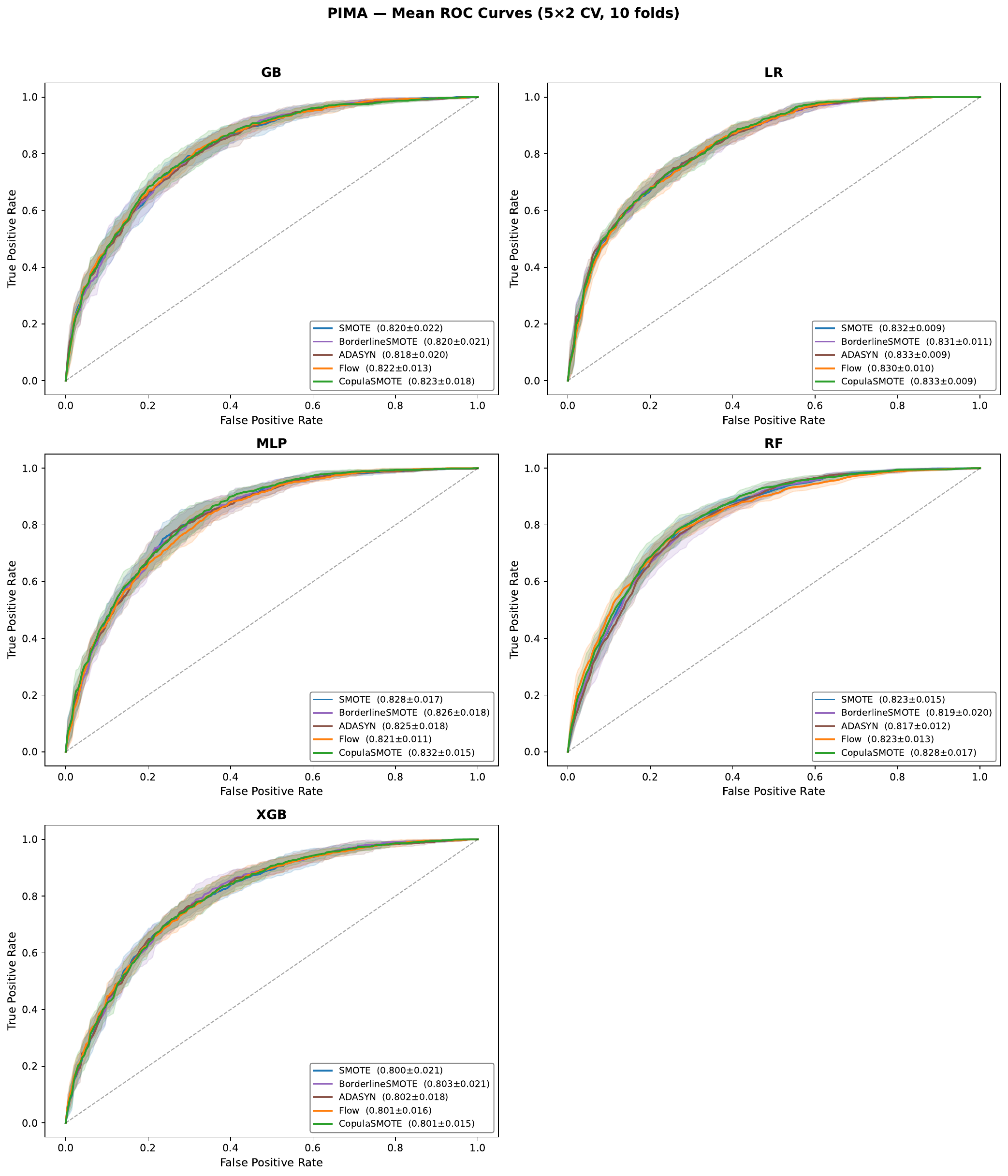}
    \caption{Mean ROC curves for the five classifiers on the Pima Indians
    Diabetes dataset, averaged across the ten folds of the 5$\times$2
    cross-validation. Shaded bands indicate $\pm 1$ standard deviation.
    Legend entries report mean AUC $\pm$ standard deviation.}
    \label{fig:pima_roc}
\end{figure}

\begin{figure}[htbp]
    \centering
    \includegraphics[width=\linewidth]{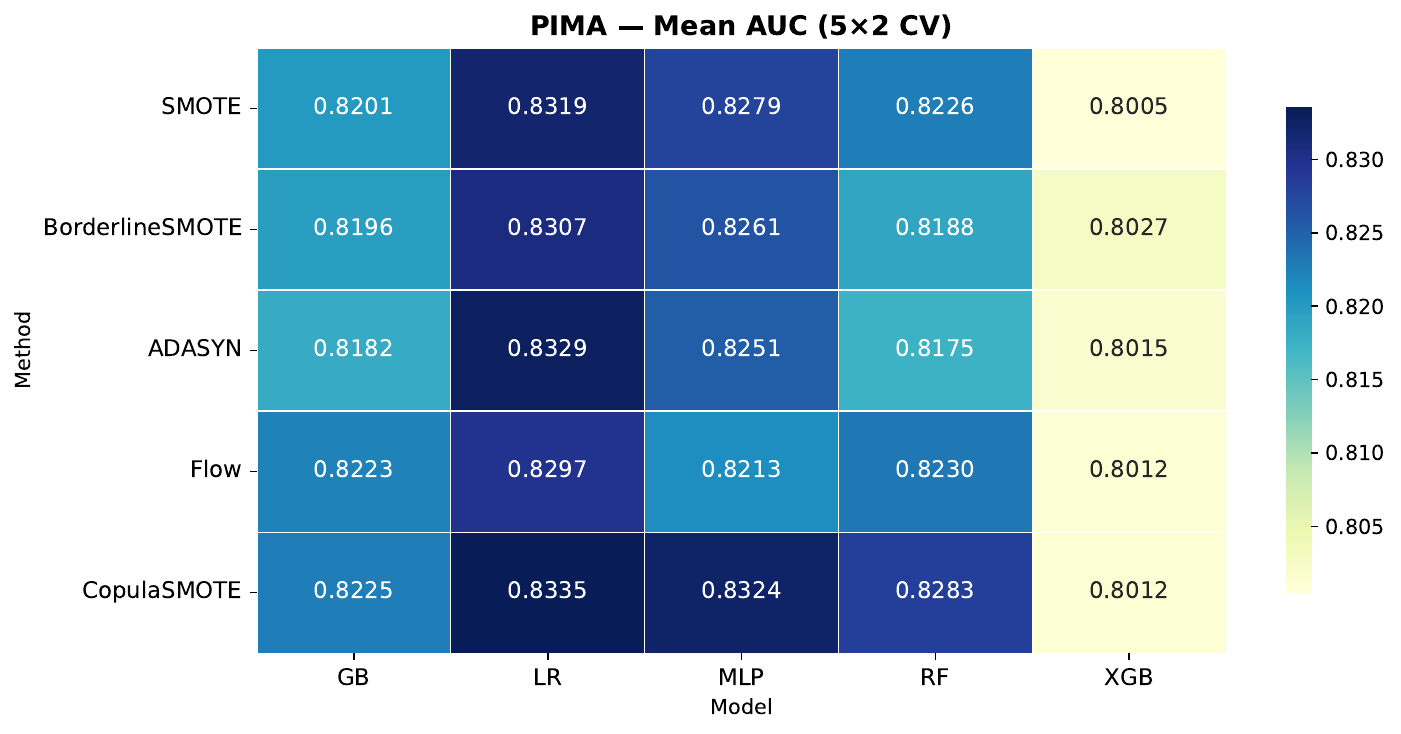}
    \caption{Mean AUC on the Pima Indians Diabetes dataset by resampling method
    and classifier, averaged across the ten folds. The narrow range across all
    cells reflects the convergent behaviour of all methods in this
    low-dimensional setting.}
    \label{fig:pima_heatmap}
\end{figure}

The ROC curves in Figure~\ref{fig:pima_roc} show substantial overlap among
the resampling methods. Similarly, the heatmap in Figure~\ref{fig:pima_heatmap}
shows that AUC values vary only slightly across classifiers and resampling
strategies.

These findings are consistent with the expected behavior of the method in a
small, low-dimensional setting. The Pima dataset contains fewer than 270
minority-class observations and only eight predictors, which limits the amount
of dependence structure that a vine copula can estimate reliably. Therefore,
CopulaSMOTE remains competitive, but the evidence does not support a strong
performance advantage over interpolation-based methods on this dataset.
\begin{table}[ht]
\centering
\caption{Classification performance on the Pima Indians Diabetes and Iraqi
Diabetes datasets: mean $\pm$ standard deviation across the ten folds of the
5$\times$2 cross-validation. Best value within each classifier column is shown
in bold. Iraqi values are subject to the ceiling effect discussed in
Section~\ref{subsec:iraqi} and should be interpreted alongside the
minority-class metrics reported in the text.}
\label{tab:pima_iraqi_results}
\resizebox{\textwidth}{!}{%
\begin{tabular}{llccccc}
\toprule
Metric & Method & RF & GB & XGB & LR & MLP \\
\midrule
\multicolumn{7}{l}{\textbf{Panel A\;:\;Pima Indians Diabetes}
  \;($n=768$,\;$d=8$,\;minority rate 34.9\%)} \\[2pt]
\multicolumn{7}{l}{\textit{F1 score}} \\
& SMOTE            & $0.6618 \pm 0.0208$ & $0.6570 \pm 0.0303$ & $0.6340 \pm 0.0216$ & $0.6612 \pm 0.0215$ & $0.6674 \pm 0.0301$ \\
& Borderline-SMOTE & $0.6575 \pm 0.0270$ & $0.6596 \pm 0.0264$ & $0.6320 \pm 0.0268$ & $\mathbf{0.6687 \pm 0.0207}$ & $0.6704 \pm 0.0304$ \\
& ADASYN           & $0.6598 \pm 0.0178$ & $0.6548 \pm 0.0237$ & $0.6393 \pm 0.0221$ & $0.6665 \pm 0.0154$ & $\mathbf{0.6744 \pm 0.0272}$ \\
& Flow             & $0.6651 \pm 0.0153$ & $0.6463 \pm 0.0197$ & $0.6334 \pm 0.0267$ & $0.6588 \pm 0.0238$ & $0.6491 \pm 0.0177$ \\
& CopulaSMOTE      & $\mathbf{0.6829 \pm 0.0210}$ & $\mathbf{0.6667 \pm 0.0272}$ & $\mathbf{0.6436 \pm 0.0261}$ & $0.6621 \pm 0.0269$ & $0.6686 \pm 0.0280$ \\[4pt]
\multicolumn{7}{l}{\textit{AUC}} \\
& SMOTE            & $0.8226 \pm 0.0153$ & $0.8201 \pm 0.0231$ & $0.8005 \pm 0.0217$ & $0.8319 \pm 0.0096$ & $0.8279 \pm 0.0180$ \\
& Borderline-SMOTE & $0.8188 \pm 0.0210$ & $0.8196 \pm 0.0222$ & $\mathbf{0.8027 \pm 0.0221}$ & $0.8307 \pm 0.0114$ & $0.8261 \pm 0.0186$ \\
& ADASYN           & $0.8175 \pm 0.0132$ & $0.8182 \pm 0.0206$ & $0.8015 \pm 0.0190$ & $0.8329 \pm 0.0092$ & $0.8251 \pm 0.0195$ \\
& Flow             & $0.8230 \pm 0.0142$ & $0.8223 \pm 0.0132$ & $0.8012 \pm 0.0166$ & $0.8297 \pm 0.0106$ & $0.8213 \pm 0.0111$ \\
& CopulaSMOTE      & $\mathbf{0.8283 \pm 0.0179}$ & $\mathbf{0.8225 \pm 0.0191}$ & $0.8012 \pm 0.0161$ & $\mathbf{0.8335 \pm 0.0092}$ & $\mathbf{0.8324 \pm 0.0157}$ \\
\midrule
\multicolumn{7}{l}{\textbf{Panel B\;:\;Iraqi Diabetes}
  \;($n=947$,\;$d=11$,\;minority rate 10.9\%)} \\[2pt]
\multicolumn{7}{l}{\textit{F1 score}} \\
& SMOTE            & $\mathbf{0.9925 \pm 0.0029}$ & $\mathbf{0.9920 \pm 0.0025}$ & $\mathbf{0.9939 \pm 0.0023}$ & $\mathbf{0.9770 \pm 0.0045}$ & $\mathbf{0.9798 \pm 0.0070}$ \\
& Borderline-SMOTE & $0.9916 \pm 0.0027$ & $0.9910 \pm 0.0038$ & $0.9933 \pm 0.0022$ & $0.9751 \pm 0.0056$ & $0.9792 \pm 0.0062$ \\
& ADASYN           & $0.9924 \pm 0.0019$ & $0.9911 \pm 0.0037$ & $0.9935 \pm 0.0019$ & $0.9760 \pm 0.0049$ & $0.9794 \pm 0.0058$ \\
& Flow             & $0.9813 \pm 0.0061$ & $0.9864 \pm 0.0057$ & $0.9844 \pm 0.0076$ & $0.9638 \pm 0.0135$ & $0.9715 \pm 0.0101$ \\
& CopulaSMOTE      & $0.9904 \pm 0.0027$ & $0.9911 \pm 0.0033$ & $0.9931 \pm 0.0030$ & $0.9737 \pm 0.0060$ & $0.9766 \pm 0.0071$ \\[4pt]
\multicolumn{7}{l}{\textit{AUC}} \\
& SMOTE            & $\mathbf{0.9983 \pm 0.0006}$ & $0.9976 \pm 0.0012$ & $0.9986 \pm 0.0009$ & $0.9909 \pm 0.0044$ & $0.9790 \pm 0.0145$ \\
& Borderline-SMOTE & $0.9977 \pm 0.0011$ & $0.9967 \pm 0.0023$ & $0.9983 \pm 0.0016$ & $0.9902 \pm 0.0038$ & $0.9769 \pm 0.0173$ \\
& ADASYN           & $0.9980 \pm 0.0007$ & $0.9972 \pm 0.0014$ & $0.9986 \pm 0.0009$ & $0.9903 \pm 0.0044$ & $0.9785 \pm 0.0154$ \\
& Flow             & $0.9956 \pm 0.0021$ & $0.9958 \pm 0.0027$ & $0.9952 \pm 0.0028$ & $0.9785 \pm 0.0133$ & $0.9837 \pm 0.0072$ \\
& CopulaSMOTE      & $0.9981 \pm 0.0007$ & $\mathbf{0.9981 \pm 0.0009}$ & $\mathbf{0.9987 \pm 0.0008}$ & $\mathbf{0.9910 \pm 0.0032}$ & $\mathbf{0.9886 \pm 0.0080}$ \\
\bottomrule
\end{tabular}%
}
\end{table}

\subsection{Iraqi Diabetes Dataset}\label{subsec:iraqi}

The Iraqi dataset presents a different setting: eleven features,
947 observations after filtering, and a severe class imbalance in which the
non-diabetic group constitutes only 10.9\% of the sample. Because the diabetic
majority is comparatively easy to identify, overall performance metrics
approach saturation across all methods and classifiers.

\begin{figure}[htbp]
    \centering
    \includegraphics[width=\textwidth]{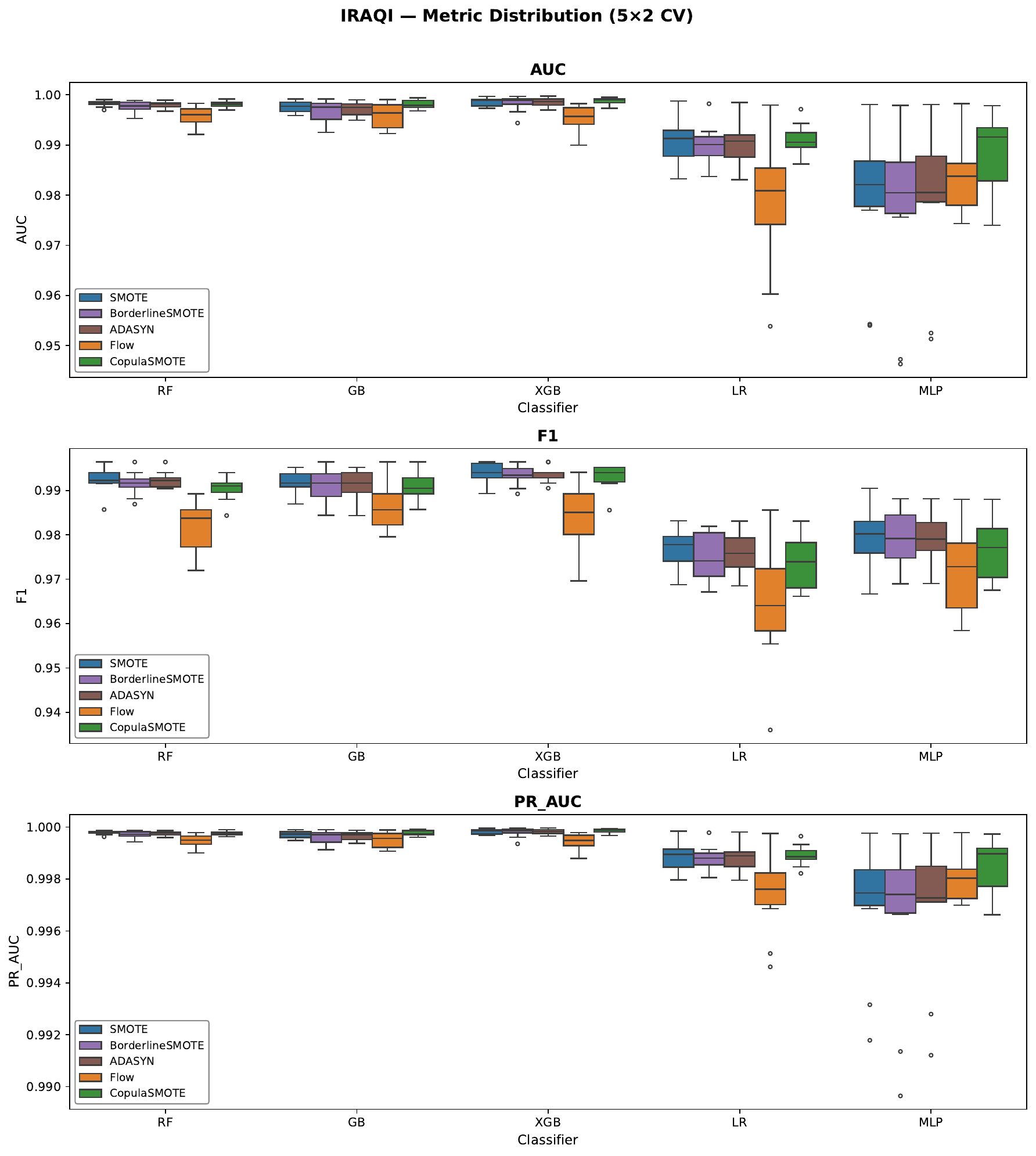}
    \caption{Distribution of AUC, F1, and PR~AUC across the ten folds on the
    Iraqi Diabetes dataset. The tight, overlapping distributions confirm that
    no method separates clearly under the ceiling effect.}
    \label{fig:iraqi_boxplots}
\end{figure}

As shown in Table~\ref{tab:pima_iraqi_results} (Panel~B), AUC values cluster
between 0.977 and 0.999 and F1 values between 0.964 and 0.994, leaving little
room for separation among methods. On overall F1, the SMOTE family performs
slightly better across most classifiers, while CopulaSMOTE attains the highest
AUC for four of the five classifiers. None of these differences are
statistically significant. This ceiling effect is also apparent in the tightly overlapping metric
distributions shown in Figure~\ref{fig:iraqi_boxplots}.

Minority-class-focused metrics provide a more informative comparison in this
setting. CopulaSMOTE with XGBoost achieves the highest minority-class recall
($0.9922 \pm 0.0101$) and the highest minority-class PR~AUC
($0.9896 \pm 0.0068$). SMOTE with XGBoost attains a slightly higher
minority-class F1 ($0.9516$ versus $0.9468$) but substantially lower recall
($0.9726$ versus $0.9922$), indicating that CopulaSMOTE recovers a larger
fraction of minority observations at the expense of a modest precision loss.

Overall, the Iraqi results suggest that CopulaSMOTE maintains competitive
performance in severely imbalanced small-sample settings, although the ceiling
effect limits the extent to which meaningful differences can emerge.


\begin{figure}[htbp]
    \centering
    \includegraphics[width=\linewidth]{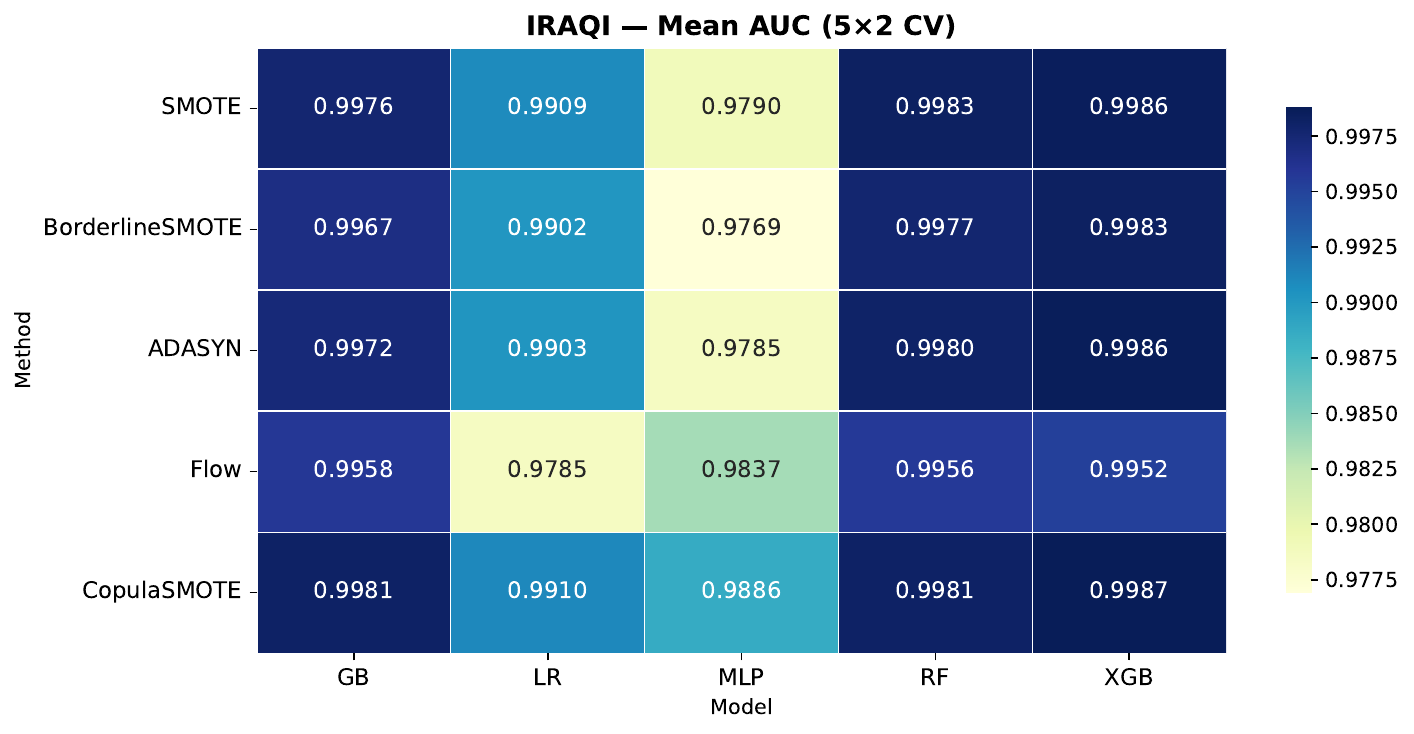}
    \caption{Mean AUC on the Iraqi Diabetes dataset by resampling method and
    classifier, averaged across the ten folds.}
    \label{fig:iraqi_heatmap}
\end{figure}

\subsection{CDC BRFSS 2015 Diabetes Health Indicators}\label{subsec:cdc}

The CDC dataset is the primary benchmark of this study. With 253,680
observations, 21 predictors, and a 13.9\% minority rate, it provides a setting
in which complex dependence structure is both estimable and potentially useful
for synthetic minority generation.

Table~\ref{tab:cdc_main_results} summarizes mean F1 and AUC across all
methods and classifiers. Three broad patterns emerge. First, CopulaSMOTE
achieves the highest F1 across all five classifiers, with particularly large
improvements for random forest and XGBoost. Second, the normalizing flow
produces the weakest F1 across all classifiers. Third, AUC results are more
mixed: CopulaSMOTE performs best for logistic regression, while the
normalizing flow attains the highest AUC for gradient boosting, XGBoost,
and MLP.

\begin{figure}[htbp]
    \centering
    \includegraphics[width=\linewidth]{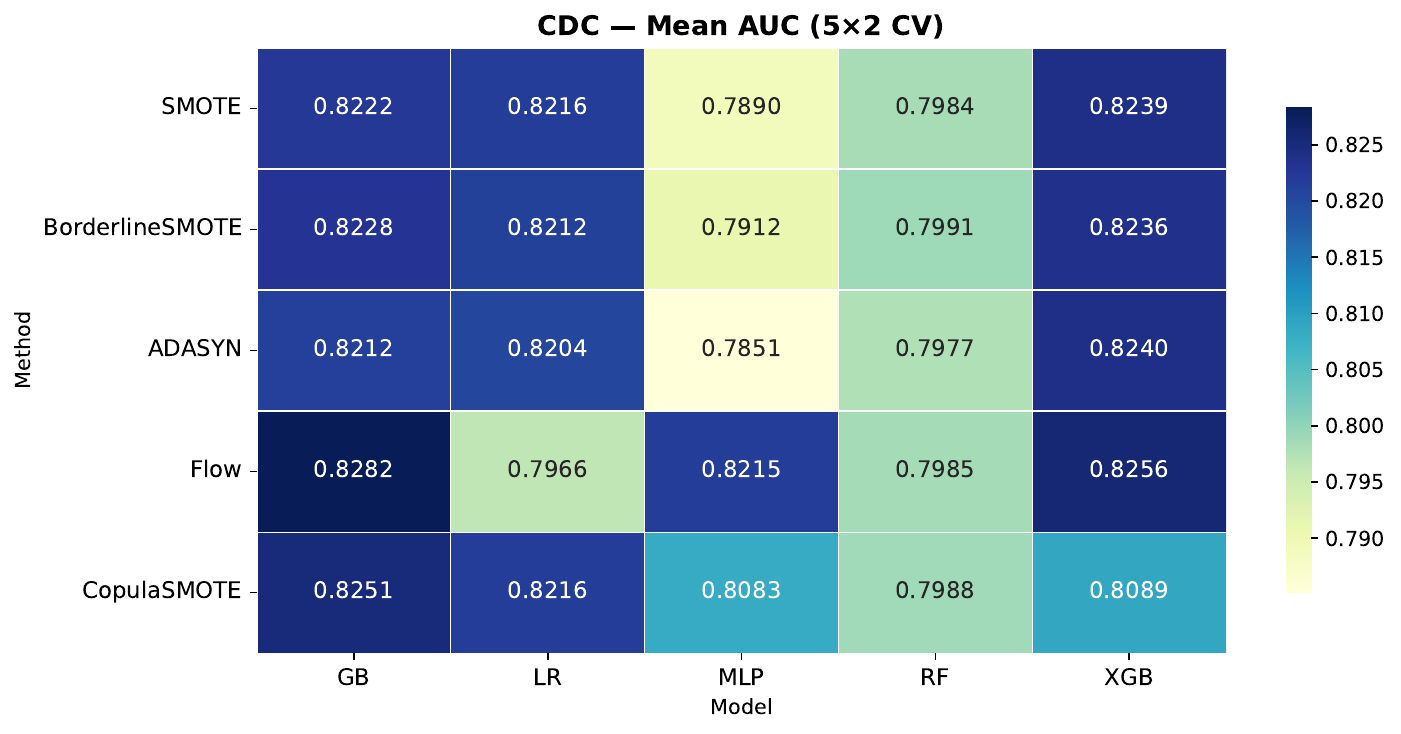}
    \caption{Mean AUC on the CDC dataset by resampling method and
    classifier, averaged across the ten folds. CopulaSMOTE leads for logistic
    regression; the normalizing flow leads for gradient boosting, XGBoost,
    and MLP.}
    \label{fig:cdc_heatmap}
\end{figure}

These patterns are also visible in the CDC heatmap
(Figure~\ref{fig:cdc_heatmap}), where CopulaSMOTE dominates on F1 while
the normalizing flow remains competitive on AUC for several classifiers.

\begin{table}[htbp]
\centering
\caption{Classification performance on the CDC dataset: mean $\pm$ standard deviation across the ten folds of the
5$\times$2 cross-validation. Best value within each classifier column is shown
in bold.}
\label{tab:cdc_main_results}
\resizebox{\textwidth}{!}{%
\begin{tabular}{llccccc}
\toprule
Metric & Method & RF & GB & XGB & LR & MLP \\
\midrule
\multirow{5}{*}{F1}
& SMOTE            & $0.3428 \pm 0.0022$ & $0.4383 \pm 0.0034$ & $0.2854 \pm 0.0033$ & $0.4427 \pm 0.0015$ & $0.4219 \pm 0.0057$ \\
& Borderline-SMOTE & $0.3392 \pm 0.0028$ & $0.4442 \pm 0.0028$ & $0.2802 \pm 0.0036$ & $0.4440 \pm 0.0018$ & $0.4213 \pm 0.0094$ \\
& ADASYN           & $0.3349 \pm 0.0026$ & $0.4332 \pm 0.0031$ & $0.2771 \pm 0.0033$ & $0.4377 \pm 0.0012$ & $0.4166 \pm 0.0076$ \\
& Flow             & $0.2494 \pm 0.0054$ & $0.2717 \pm 0.0057$ & $0.2628 \pm 0.0039$ & $0.4246 \pm 0.0124$ & $0.2889 \pm 0.0285$ \\
& CopulaSMOTE      & $\mathbf{0.4260 \pm 0.0019}$ & $\mathbf{0.4453 \pm 0.0014}$ & $\mathbf{0.4357 \pm 0.0015}$ & $\mathbf{0.4455 \pm 0.0019}$ & $\mathbf{0.4336 \pm 0.0025}$ \\
\midrule
\multirow{5}{*}{AUC}
& SMOTE            & $0.7984 \pm 0.0012$ & $0.8222 \pm 0.0019$ & $0.8239 \pm 0.0020$ & $0.8216 \pm 0.0019$ & $0.7890 \pm 0.0045$ \\
& Borderline-SMOTE & $\mathbf{0.7991 \pm 0.0013}$ & $0.8228 \pm 0.0019$ & $0.8236 \pm 0.0014$ & $0.8212 \pm 0.0020$ & $0.7912 \pm 0.0076$ \\
& ADASYN           & $0.7977 \pm 0.0014$ & $0.8212 \pm 0.0021$ & $0.8240 \pm 0.0015$ & $0.8204 \pm 0.0018$ & $0.7851 \pm 0.0071$ \\
& Flow             & $0.7985 \pm 0.0013$ & $\mathbf{0.8282 \pm 0.0018}$ & $\mathbf{0.8256 \pm 0.0015}$ & $0.7966 \pm 0.0117$ & $\mathbf{0.8215 \pm 0.0016}$ \\
& CopulaSMOTE      & $0.7988 \pm 0.0020$ & $0.8251 \pm 0.0020$ & $0.8089 \pm 0.0021$ & $\mathbf{0.8216 \pm 0.0019}$ & $0.8083 \pm 0.0027$ \\
\bottomrule
\end{tabular}%
}
\end{table}

\paragraph{F1 score.}
The largest improvements occur for random forest and XGBoost relative to the
SMOTE family. CopulaSMOTE exceeds SMOTE by $+0.0832$ F1 points for random
forest and by $+0.1503$ for XGBoost, both significant at $p < 0.0001$.
Differences relative to Borderline-SMOTE and ADASYN are of similar magnitude.
Against the normalizing flow, CopulaSMOTE improves F1 by more than 0.14 points
for every classifier except logistic regression.

\paragraph{AUC.}
The AUC results are more heterogeneous. CopulaSMOTE improves significantly
over the SMOTE family for gradient boosting and MLP, but performs
significantly worse for XGBoost, where its AUC
($0.8089 \pm 0.0021$) falls below all baselines
($p < 0.001$ versus each SMOTE variant and the flow). For random forest and
logistic regression, CopulaSMOTE remains statistically comparable to the
SMOTE family. The normalizing flow achieves the highest AUC for gradient
boosting, XGBoost, and MLP.

\paragraph{Balanced accuracy and minority-class recall.}
CopulaSMOTE with gradient boosting achieves the highest balanced accuracy
among all 25 method-classifier combinations
($0.7476 \pm 0.0020$). The associated minority-class recall
($0.7669 \pm 0.0063$) substantially exceeds that of SMOTE
($0.4461 \pm 0.0093$) and Borderline-SMOTE
($0.4660 \pm 0.0068$) with the same classifier. This pattern indicates that
the vine-generated synthetic observations move the decision boundary more
aggressively toward the minority class than interpolation-based methods.

\paragraph{PR AUC.}
CopulaSMOTE underperforms the normalizing flow on PR~AUC for gradient
boosting, MLP, and XGBoost, and also underperforms the SMOTE family for
XGBoost. Conversely, CopulaSMOTE achieves higher PR~AUC for logistic
regression versus ADASYN and for MLP versus all three SMOTE variants.
The lower PR~AUC observed for XGBoost is consistent with the recall-oriented
behaviour noted above: higher minority-class coverage is accompanied by lower
precision at operating points emphasized by PR~AUC.

\paragraph{Significance summary for CDC.}
Table~\ref{tab:cdc_significance} summarizes representative statistically
significant comparisons. Across all 60 pairwise Dietterich tests,
CopulaSMOTE is significantly better in 25 cases and significantly worse in 12.
The gains are concentrated in F1 comparisons against the SMOTE family and the
normalizing flow, while the losses are concentrated in AUC and PR~AUC for
XGBoost.

The XGBoost results illustrate the central tradeoff observed throughout the
CDC experiments. The vine-generated synthetic observations shift the classifier
toward higher minority-class recall, substantially improving F1 while reducing
AUC and PR~AUC due to lower precision at low recall thresholds. Gradient
boosting, logistic regression, and MLP appear more robust to this shift and
benefit accordingly.

\begin{figure}[htbp]
    \centering
    \includegraphics[width=\textwidth,height=0.8\textheight,keepaspectratio]{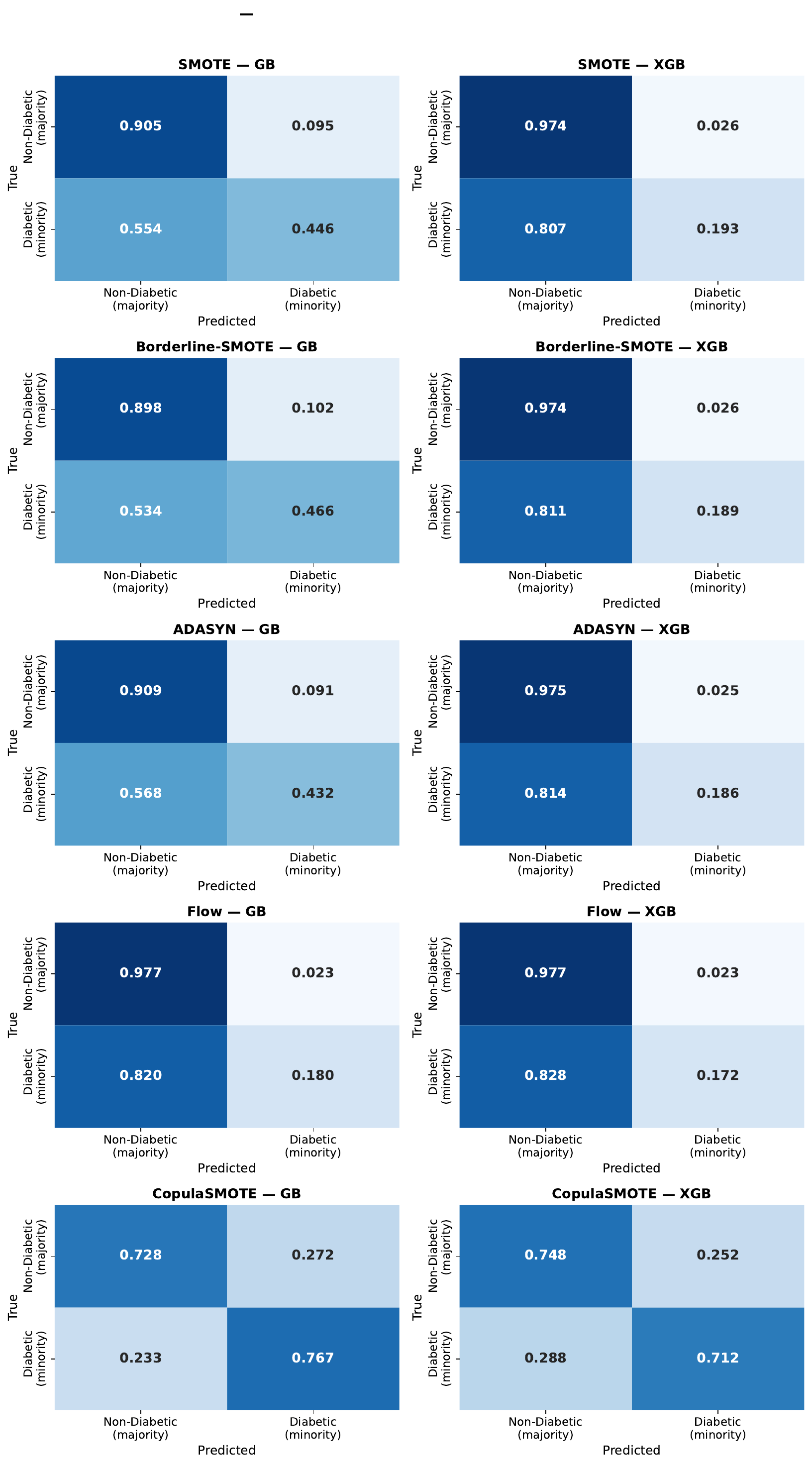}
   \caption{Mean row-normalised confusion matrices for gradient boosting (GB)
and XGBoost (XGB) on the CDC dataset, averaged across the ten
folds. GB illustrates CopulaSMOTE's higher minority-class recall; XGB
illustrates the associated false-positive cost.}
    \label{fig:cdc_confusion}
\end{figure}

The confusion matrices in Figure~\ref{fig:cdc_confusion} provide a visual
summary of this behaviour, showing the increase in minority-class recovery
together with the associated rise in false positives.

\begin{table*}[htbp]
\centering
\caption{Selected statistically significant pairwise comparisons between
CopulaSMOTE and baseline methods on the CDC dataset (Dietterich
5$\times$2 paired $t$-test, $\alpha = 0.05$). Positive differences favour
CopulaSMOTE; negative differences favour the baseline.}
\label{tab:cdc_significance}
\begin{tabular}{lllcccc}
\toprule
Classifier & Metric & Comparison & CopulaSMOTE & Baseline & Difference & $p$-value \\
\midrule
\multicolumn{7}{l}{\textit{CopulaSMOTE outperforms SMOTE family on F1}} \\
RF  & F1 & vs SMOTE            & 0.4260 & 0.3428 & $+0.0832$ & $<0.0001$ \\
RF  & F1 & vs Borderline-SMOTE & 0.4260 & 0.3392 & $+0.0868$ & $<0.0001$ \\
RF  & F1 & vs ADASYN           & 0.4260 & 0.3349 & $+0.0911$ & $<0.0001$ \\
XGB & F1 & vs SMOTE            & 0.4357 & 0.2854 & $+0.1503$ & $<0.0001$ \\
XGB & F1 & vs Borderline-SMOTE & 0.4357 & 0.2802 & $+0.1555$ & $<0.0001$ \\
XGB & F1 & vs ADASYN           & 0.4357 & 0.2771 & $+0.1586$ & $<0.0001$ \\
RF  & F1 & vs Flow             & 0.4260 & 0.2494 & $+0.1766$ & $<0.0001$ \\
\midrule
\multicolumn{7}{l}{\textit{CopulaSMOTE outperforms SMOTE family on AUC}} \\
GB  & AUC & vs SMOTE           & 0.8251 & 0.8222 & $+0.0029$ & $0.022$ \\
GB  & AUC & vs Borderline-SMOTE& 0.8251 & 0.8228 & $+0.0023$ & $0.035$ \\
GB  & AUC & vs ADASYN          & 0.8251 & 0.8212 & $+0.0039$ & $0.029$ \\
MLP & AUC & vs SMOTE           & 0.8083 & 0.7890 & $+0.0192$ & $0.001$ \\
MLP & AUC & vs Borderline-SMOTE& 0.8083 & 0.7912 & $+0.0171$ & $0.008$ \\
MLP & AUC & vs ADASYN          & 0.8083 & 0.7851 & $+0.0231$ & $0.018$ \\
\midrule
\multicolumn{7}{l}{\textit{Baselines outperform CopulaSMOTE}} \\
XGB & AUC    & vs SMOTE        & 0.8089 & 0.8239 & $-0.0150$ & $0.001$ \\
XGB & PR AUC & vs SMOTE        & 0.3696 & 0.4200 & $-0.0505$ & $<0.001$ \\
XGB & PR AUC & vs ADASYN       & 0.3696 & 0.4203 & $-0.0507$ & $<0.001$ \\
GB  & AUC    & vs Flow         & 0.8251 & 0.8282 & $-0.0031$ & $0.034$ \\
RF  & PR AUC & vs Flow         & 0.3565 & 0.3712 & $-0.0147$ & $0.032$ \\
MLP & PR AUC & vs Flow         & 0.3677 & 0.4104 & $-0.0427$ & $0.009$ \\
\bottomrule
\end{tabular}
\end{table*}

\begin{figure}[htbp]
    \centering
    \includegraphics[width=\textwidth]{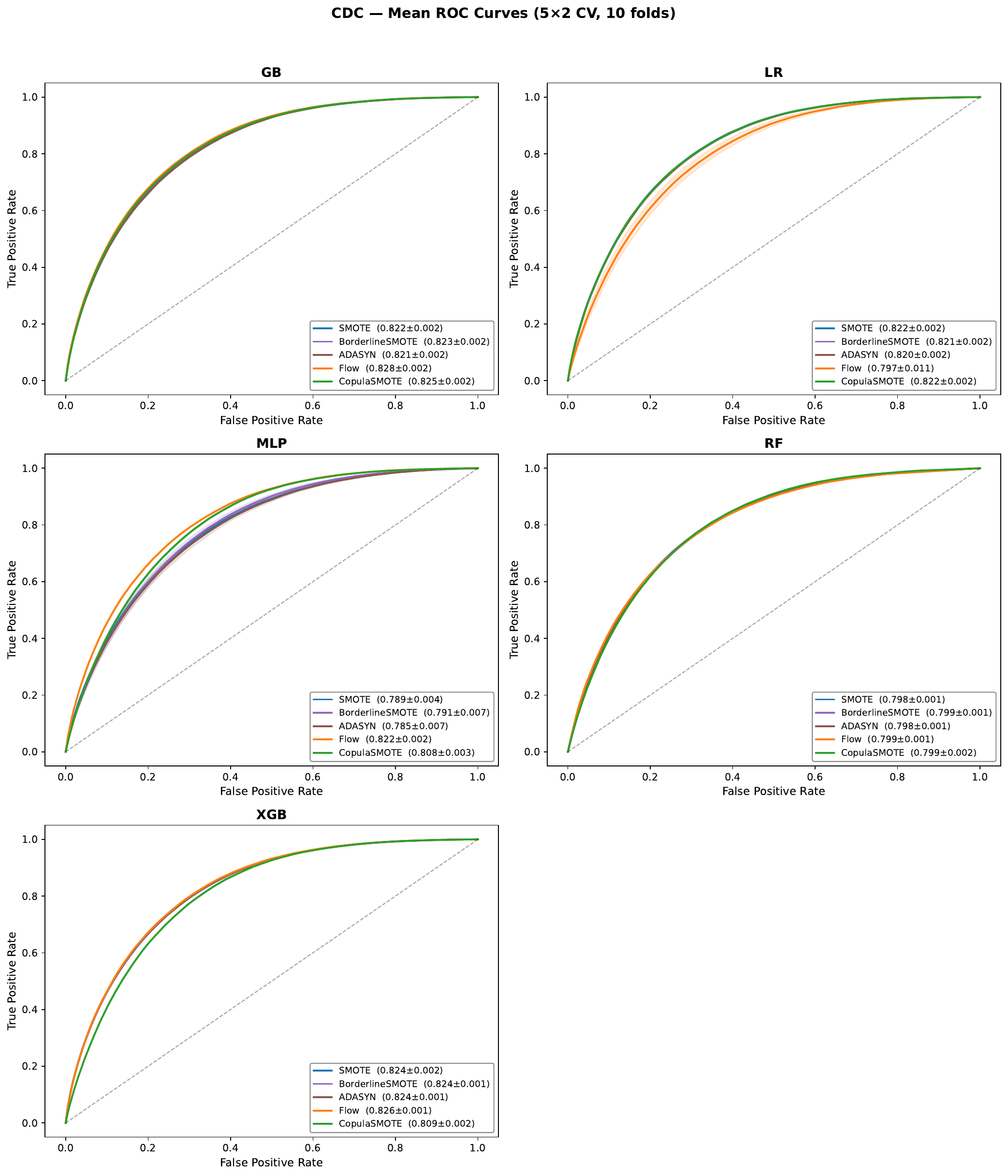}
    \caption{Mean ROC curves for the five classifiers on the CDC
    dataset, averaged across the ten folds. Shaded bands indicate $\pm 1$
    standard deviation. Legend entries report mean AUC $\pm$ standard
    deviation.}
    \label{fig:cdc_roc}
\end{figure}

\begin{figure}[htbp]
    \centering
    \includegraphics[width=\linewidth]{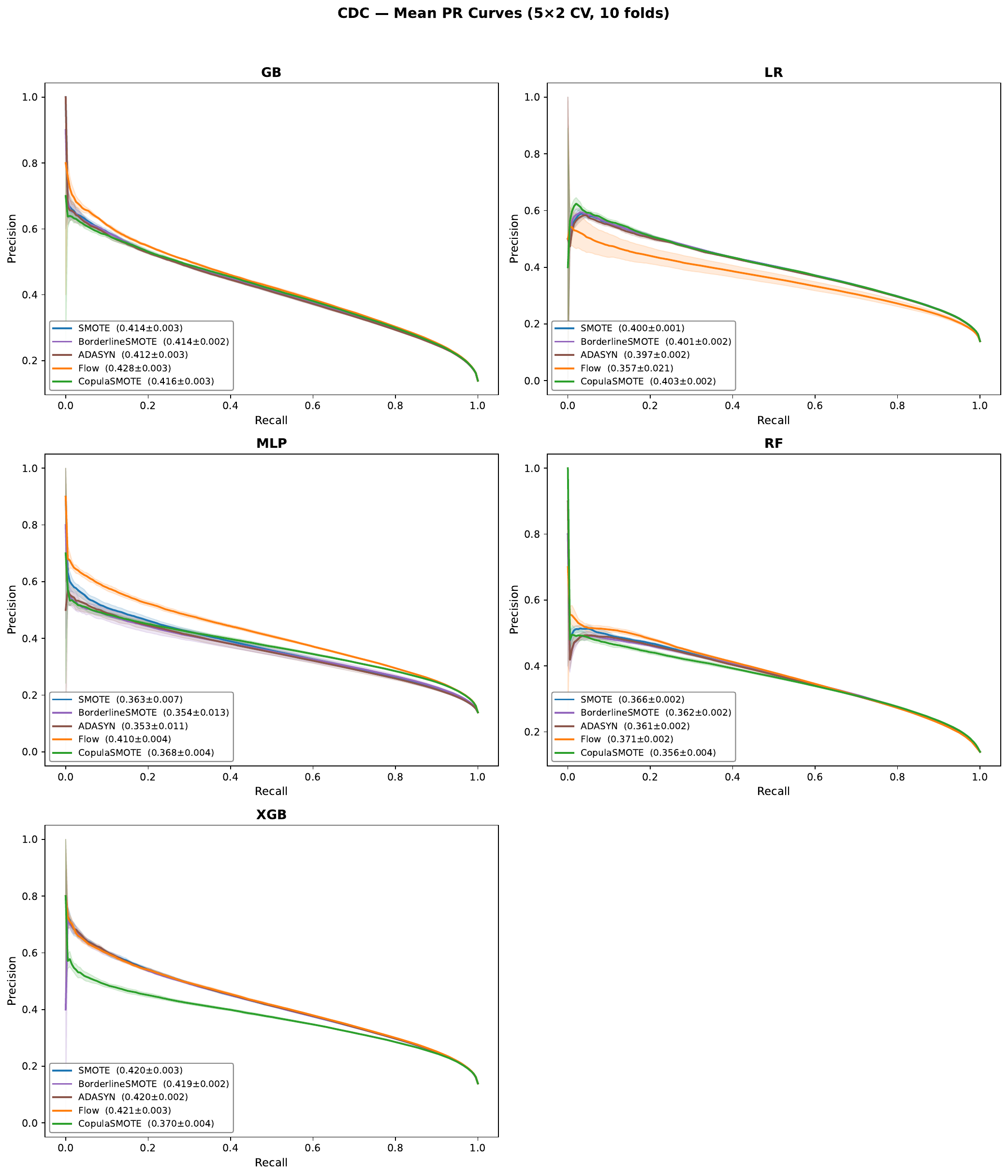}
    \caption{Mean precision--recall curves for the five classifiers on the CDC
    BRFSS 2017 dataset, averaged across the ten folds. The XGBoost panel
    illustrates the precision--recall tradeoff: CopulaSMOTE achieves higher
    recall but lower precision at low recall thresholds relative to all
    baselines.}
    \label{fig:cdc_pr}
\end{figure}


\subsection{Cross-Dataset Patterns}\label{subsec:patterns}

Table~\ref{tab:significance_summary} aggregates the Dietterich test results
across all three datasets and summarizes how CopulaSMOTE's performance changes
with dataset scale and dimensionality.

\begin{table}[htbp]
\centering
\caption{Summary of statistically significant pairwise Dietterich 5$\times$2
$t$-test comparisons ($\alpha = 0.05$) between CopulaSMOTE and the four
baselines combined, per metric and dataset. W\,=\,number of comparisons where
CopulaSMOTE is significantly better; L\,=\,number where it is significantly
worse. Each metric yields 20 possible pairwise comparisons
(5 classifiers $\times$ 4 baselines); the Total column covers all three metrics.}
\label{tab:significance_summary}
\begin{tabular}{lcccccccc}
\toprule
& \multicolumn{2}{c}{F1} & \multicolumn{2}{c}{AUC} &
  \multicolumn{2}{c}{PR AUC} & \multicolumn{2}{c}{Total} \\
\cmidrule(lr){2-3}\cmidrule(lr){4-5}\cmidrule(lr){6-7}\cmidrule(lr){8-9}
Dataset & W & L & W & L & W & L & W & L \\
\midrule
Pima  \;($n=768$,\;$d=8$)  & 1 & 0 & 0 & 0 & 0 & 0 &  1 &  0 \\
Iraqi \;($n=947$,\;$d=11$) & 0 & 0 & 0 & 0 & 0 & 0 &  0 &  0 \\
CDC   \;($n=253{,}680$,\;$d=21$) & 16 & 0 & 6 & 5 & 3 & 7 & 25 & 12 \\
\bottomrule
\end{tabular}
\end{table}

Across datasets, the ROC, PR, and heatmap visualizations consistently mirror
the statistical comparisons reported in the tables, with the clearest visual
separation appearing on the CDC dataset.

Three patterns emerge from the combined results.

On the Pima and Iraqi datasets, which contain at most eleven features and fewer
than 270 minority observations, few statistically significant differences are
detected. In contrast, the CDC dataset contains 21 predictors and more than
35,000 minority observations, and CopulaSMOTE records 25 significant wins
across 60 comparisons. This pattern suggests that dependence-aware synthetic
generation becomes more effective as dimensional structure and minority sample
size increase.

A second pattern is the consistent shift toward higher minority-class recall
relative to interpolation-based methods. Across datasets and classifiers,
CopulaSMOTE frequently improves F1, particularly on the CDC dataset, but this
gain is sometimes accompanied by reductions in AUC and PR~AUC when precision
losses become substantial.

The normalizing flow exhibits a different behaviour. Although it achieves
competitive or superior AUC in several cases, especially on the CDC dataset,
it consistently produces weaker F1 scores. This pattern is consistent with the
difficulty of fitting a stable high-capacity generative density model when
minority sample sizes remain relatively small.

Taken together, the results position CopulaSMOTE as a practical complement
to the SMOTE family for high-dimensional clinical survey data, with the
clearest advantages appearing when the minority class is sufficiently large
to support reliable vine copula estimation.

\section{Conclusion and Future Work}\label{sec6}

We have introduced CopulaSMOTE, a vine copula-based oversampling framework
for imbalanced clinical classification. The method fits a truncated vine
copula to the rank-transformed minority class, simulates new pseudo
observations from the fitted vine, and maps them back to the original feature
space through empirical inverse cumulative distribution functions. Relative
to SMOTE and its variants, the method models the joint dependence structure
of the minority class as an explicit object rather than as a byproduct of
local geometric interpolation.

The empirical results across three publicly available diabetes datasets reveal
a consistent and interpretable pattern. On the Pima and Iraqi datasets, where
the minority class contains fewer than 270 observations and at most eleven
features, CopulaSMOTE remains competitive with the strongest baselines but
produces few statistically significant differences in either direction. On the
CDC BRFSS dataset, with 21 features and more than 35,000 minority observations,
CopulaSMOTE records 25 significant wins and 12 losses across 60 pairwise
Dietterich tests. Wins concentrate in F1 for all five classifiers against the
SMOTE family and the normalizing flow. Losses concentrate in AUC and PR~AUC
for XGBoost, where the vine-generated samples shift the classifier toward
higher recall at the cost of lower precision, depressing metrics that
penalize precision losses at low recall thresholds. The normalizing flow
achieves competitive or superior AUC on the CDC dataset for several
classifiers but consistently produces the weakest F1 across all three
datasets, reflecting the difficulty of fitting a stable high-capacity
generative model when minority sample sizes are modest. Therefore, CopulaSMOTE should be viewed as a practical tool for recall-oriented clinical
screening tasks rather than as a uniformly dominant replacement for existing oversampling methods.
Its strongest use case appears to be moderate-dimensional tabular data with enough minority-class
observations to estimate a stable vine-copula dependence structure.


Several directions are worth pursuing in future work. A natural first
extension is to replace the continuous extension jitter used here with a
fully specified mixed data vine copula model in which discrete and continuous
features are handled under a unified likelihood, building on recent
developments in D-vine quantile regression with discrete variables
\cite{schallhorn2017d} and mixed margin vine copula estimation
\cite{nagler2018vine}. A second direction is to evaluate CopulaSMOTE on
clinical datasets with multiclass or time-to-event outcomes, where the
notion of a minority class takes on a different form. A third direction is
to study subgroup-level behavior, assessing whether the synthetic
observations preserve dependence structure within demographic subgroups, which
is relevant when clinical prediction fairness is a concern. A fourth
direction is to examine the sensitivity of results to the vine truncation
level and the bivariate family library, both of which were held fixed here
to keep the comparison focused on the oversampling method itself.

Finally, the current framework generates synthetic observations per fold
within a cross-validation protocol. Extending it to external validation and
transport settings raises questions about whether to refit the vine copula
at the deployment site or carry over the development-site model, each of
which has implications for calibration and transportability that warrant
separate investigation.


\section*{Data Availability}

The datasets used in this study are publicly available from online repositories. The PIMA Indians Diabetes dataset was obtained from the Kaggle Machine Learning Repository (\url{https://www.kaggle.com/datasets/uciml/pima-indians-diabetes-database}). The Iraqi Diabetes dataset was also sourced from Kaggle (\url{https://www.kaggle.com/datasets/aravindpcoder/diabetes-dataset}). In addition, the CDC BRFSS 2015 Diabetes Health Indicators dataset was obtained from Kaggle (\url{https://www.kaggle.com/datasets/alexteboul/diabetes-health-indicators-dataset/data}).

\section*{Code Availability}

All code used in this work is available at: \url{https://github.com/agnivibes/copulasmote-diabetes-classification}.

\section*{Author Contributions}
A.A. developed and implemented the full computational pipeline, including the vine copula-based oversampling framework and all baseline methods, and selected the machine learning components. M.M.M. proposed the original idea of using copula-based methods in this study and contributed to the Introduction, Related Work, and reference development. B.W. provided guidance on model selection and experimental design. S.H. contributed to the coding, selected the datasets, wrote the final version of the manuscript, and provided overall supervision of the study. All authors reviewed and approved the final manuscript.


\end{document}